\documentclass[conference]{IEEEtran}
\IEEEoverridecommandlockouts                              

\usepackage[pdftex]{graphicx}
\usepackage{cite}
\usepackage{hyperref}
\usepackage{import}
\usepackage{xcolor}
\usepackage{amsmath}
\usepackage{amssymb}
\usepackage{fontawesome}
\usepackage{colortbl}
\usepackage{caption}
\usepackage{multirow}
\usepackage{cases}
\usepackage{booktabs}
\usepackage{algorithm}
\usepackage{algpseudocode}
\usepackage{siunitx}
\usepackage{arydshln}
\usepackage{varwidth}

\usepackage{pgf}
\usepackage{tikz}
\usetikzlibrary{arrows,automata}
\usetikzlibrary{positioning}
\tikzset{    state2/.style={ rectangle,  draw=black, inner sep=7pt,  text centered  },}
\pgfkeys{/pgf/transparency group/.code={}}

\title{\LARGE \bf
Risk-Based Filtering of Valuable Driving Situations \\ in the Waymo Open Motion Dataset
}

\author{
\IEEEauthorblockN{Tim Puphal*}  
\IEEEauthorblockA{\textit{Honda Research Institute Europe} \\  
\textit{Honda Research Institute Japan} \\  
\texttt{tim.puphal@honda-ri.de} \\
*Corresponding author \\
\vspace*{-1.0cm}}
\and  
\IEEEauthorblockN{Vipul Ramtekkar}  
\IEEEauthorblockA{\textit{Honda R{\&}D Co., Ltd.} \\  
351-0114 Saitama, Japan \\  
\texttt{ramtekkar\_vipul@jp.honda}} 
\and  
\IEEEauthorblockN{Kenji Nishimiya}  
\IEEEauthorblockA{\textit{Honda R{\&}D Co., Ltd.} \\  
351-0114 Saitama, Japan \\  
\texttt{kenji\_nishimiya@jp.honda}
}  
}

\begin{document}

\maketitle
\thispagestyle{empty}
\pagestyle{empty}

\begin{abstract}
Improving automated vehicle software requires driving data rich in valuable road user interactions. In this paper, we propose a risk-based filtering approach that helps identify such valuable driving situations from large datasets. Specifically, we use a probabilistic risk model to detect high-risk situations. Our method stands out by considering a) first-order situations (where one vehicle directly influences another and induces risk) and b) second-order situations (where influence propagates through an intermediary vehicle). In experiments, we show that our approach effectively selects valuable driving situations in the Waymo Open Motion Dataset.  Compared to the two baseline interaction metrics of Kalman difficulty and Tracks-To-Predict (TTP), our filtering approach identifies complex and complementary situations, enriching the quality in automated vehicle testing. The risk data is open-source: \url{https://github.com/HRI-EU/RiskBasedFiltering}.
\end{abstract}


\vspace{-0.07cm}
\section{Introduction}
Improving the software for automated vehicles requires driving data with rich and valuable interactions among road users. Waymo, NuTonomy and Argo AI have recently released datasets with hundreds of hours of driving data~\cite{waymo2021, nuscenes2020, argoverse2021}. These datasets are typically filtered using simple rules, e.g., ``agent A changed lanes'' or ``agent B crossed paths with agent~C''. In general, this method helps in finding notable driving situations. However, the filtering methods are often complex and not well explained. Moreover, the way filtering is used varies between datasets, making it difficult to compare filtered driving situations.

In this paper, we therefore propose a unified risk-based filtering approach to identify valuable driving situations from large datasets. Specifically, we use a probabilistic risk model that detects high-risk situations by modeling future collision risks \cite{puphal2019}. Our method stands out because it clearly seperates two types of interactions: a) first-order situations, where one vehicle directly influences another and induces risk and b) second-order situations, where the influence is passed through an intermediary vehicle. 

\begin{figure}[t!]
  \centering
  \vspace*{0.15cm}
  \resizebox{0.98\linewidth}{!}{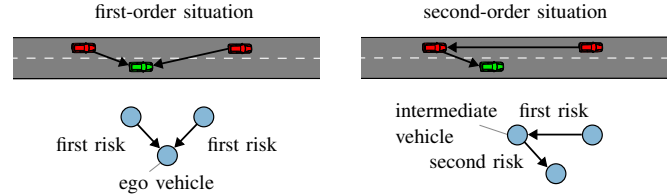}
  \vspace*{0.05cm}
  \caption[]{First-order driving situation (left) and second-order driving situation (right). In this paper, we use a probabilistic risk model to filter and identify such driving situations.} 
  \vspace*{-0.05cm}
  \label{fig:first_and_second_order_situations}
\end{figure}

Fig. \ref{fig:first_and_second_order_situations} explains in detail the two interaction types considered by the risk model. The figure shows a green ego car passing two other red cars in an adjacent lane. On the left side, the red cars directly affect the green car and create risk. This interaction is visualized as a graph, with arrows representing risk pointing directly towards the ego vehicle's node. These situations are called first-order situations. On the right side, the interaction is more indirect and analyzed differently. The front red car affects the red car behind it, which then influences the green ego car. In the graph, this is depicted by arrows showing the risk flow of from the first red car to the intermediate vehicle, and finally to the ego vehicle. These situations are called second-order situations. The risk is passed through an intermediate vehicle (i.e., first risk and second risk).

In the experiments of this paper, we apply the risk-based filtering on the Waymo Open Motion Dataset \cite{waymo2021}. By leveraging the classification of driving situations into first-order and second-order types, we demonstrate that our approach effectively selects valuable driving situations. Compared to the two common baseline interaction metrics of Kalman difficulty and Tracks-To-Predict (TTP), our method identifies complex and complementary situations, enriching the quality in automated vehicle testing.

\subsection{Related Work}
\label{sec:rel}

In related work, finding valuable driving situations in large datasets is a growing research area. Existing methods broadly fall into three categories: approaches based on changes in vehicle behavior, approaches based on number of interactions, and risk models.

One method, for example, the Kalman difficulty \cite{unitraj2024} detects behavior changes by comparing a constant velocity prediction with the actual behavior. Similarly, Tolstaya et al. \cite{tolstaya2021} introduce an interactivity score using the KL divergence between a vehicle's predicted behavior with and without considering another vehicle. This score essentially quantifies how much one vehicle changes another's behavior.

Methods focusing on the number of interactions detect valuable situations by counting trajectory combinations. For example, the authors of \cite{mavrogiannis2022} count unique combinations using topological braids. Klingelschmitt et al.~\hspace*{-0.1cm}\cite{klingelschmitt2016} also demonstrate how to identify multi-vehicle situations that lead to interactions by combining single-vehicle maneuver models. 

Lastly, risk models are often used for safety and planning, such as in Responsibility-Sensitive Safety (RSS) \cite{RSS2017}, or non-Gaussian risk modeling \cite{MIT2022}. However, they can also be used to analyze interactions, as shown in the proposed risk models \cite{puphal2019, eggert2017}. Another example is Zhan et al. \cite{interaction2019}, who used the time-to-conflict-point difference measure to find interesting driving situations.

\subsection{Contribution}
In this paper, we build on related work and use a probabilistic collision risk model to filter valuable driving situations. Our contributions are threefold:

\begin{itemize}
 \item We present a systematic classification of first-order driving situations and second-order driving situations. This allows to analyze interactions between two and three road users.
 \item We show that the filtered situations by the risk model are complementary to other methods in the Waymo dataset.
 \item And, we make our results publicly available. The computed risk values for the dataset and the filtered driving situations can be requested online at \url{https://github.com/HRI-EU}.
\end{itemize}

The remainder of the paper is structured as follows. Section \ref{sec:planning_with_filter} introduces the risk-based filtering approach, detailing the risk modeling for driving situations and the retrieval of both first-order and second-order situations. Section \ref{sec:exp_setup} gives explanations about the experimental setup and Section \ref{sec:results} shows the results and comparisons for the valuable driving situations in the Waymo dataset. Finally, Section \ref{sec:conclusion} gives the conclusion and outlook of this paper.

\section{Retrieving Valuabe Driving Situations}
\label{sec:planning_with_filter}

The goal of this paper is to detect valuable driving situations in large datasets. We define a situation as valuable when road users are likely to encounter high collision risks, often leading to change in behavior. In this section, we will thus describe the process of risk-based filtering. 

Fig. \ref{fig:risk_model_explanation} shows our method for retrieving valuable driving situations. Starting from a data example containing the trajectories of road users (left part), we model the uncertainty of each vehicle's future motion (middle part). This enables us to estimate collision risks by computing the overlap of the trajectory uncertainties and integrating the risks over time (right part). The driving situation is now represented as a graph, with nodes representing road users and arrows representing the estimated risk values (lower right part). In this step, we finally identify first-order and second-order situations by examining how risk propagates through the interaction graph.

\subsection{Risk Model}
\label{subsec:risk_model}
 
\begin{figure}[t!]
  \centering
  \vspace*{0.15cm}
  \resizebox{1.0\linewidth}{!}{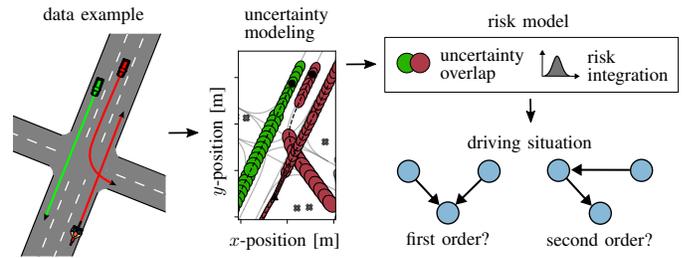}
  \vspace*{-0.5cm}
  \caption[]{Method of retrieving valuable driving situations. For each data example, the risk model is applied by analyzing the overlap of trajectory uncertainties between road users. This allows us to identify first-order and second-order situations.}
  \label{fig:risk_model_explanation}
\end{figure}

The risk model described in \cite{puphal2019} uses Gaussian distributions to model trajectory predictions and uncertainties that grow over time. We adopt the same approach. The size and growth of uncertainties aditionally depends on the vehicle type. For example, bicycles and cars have large and growing longitudinal uncertainty along their direction of motion (see Fig.~\ref{fig:risk_model_explanation}), while, e.g., pedestrians do not have large longitudinal uncertainty.

\subsubsection{Uncertainty overlap}
The collision probability at each timestep can be computed by taking the overlap of the uncertainty areas or ``probability distributions`` of two trajectories. We calculate this probability by taking the integral of the product of two Gaussian functions
\begin{equation}\label{equ:gaus_overlap} 
    P_{\text{coll},i}=\int_{\infty}f_{\text{ego}}(x)f_{\text{other},i}(x)dx,
\end{equation}
where $f_{\text{ego}}(x)$ is the Gaussian function for the ego vehicle over possible positions $x$, and $f_{\text{other},i}(x)$ is the Gaussian function for another vehicle $i$. The Gaussian function $f$ is defined by a mean position $\boldsymbol{\mu} = [x, y]$ and variance $\mathbf{\Sigma}^2  = [\sigma_{x}^2, \sigma_{y}^2]$. 

\subsubsection{Risk integration}
The survival function \cite{puphal2019} is now used to integrate collision risk over time. We compute the total collision probability $P_{\text{coll}}$ by summing the collision probabilities between the ego vehicle and other road users. The survival function then assigns a time-dependent weight to the collision probability at a future time $t+s$ in the trajectory, starting from the current time $t$ 
\begin{equation}\label{eq:survival_function}
    S(s; t) = \exp\{-\int_{t}^{t+s}(\tau_0^{-1} + \frac{P_{\text{coll}}(s;t)}{\Delta t})ds\}.
\end{equation}
The parameters $\tau_0^{-1}$ represent here the event rate to avoid a collision, and $\Delta t$ is the time window of a collision event. The final collision risk with another road user $i$ is finally computed by integrating the weighted collision probability 
\begin{equation}
    R_i(t) = \int_{0}^{s_{\text{max}}}S(s;t)\frac{P_{\text{coll},i}(s;t)}{\Delta t}ds,
\end{equation}
with the integration maximum $s_{\text{max}}$ as the fixed prediction horizon of the trajectories.

\subsection{First-Order Driving Situations}
\label{subsec:first_order}
After calculating the risks and creating a corresponding graph, valuable driving situations can be retrieved. Algorithm \ref{alg:algo_first_order} describes the process of retrieving valuable first-order driving situations. For each data example, we iterate over all vehicles and select an ego vehicle $V_{\text{ego}}$, while considering each other vehicle as a potential interacting vehicle $V_{\text{first}}$. We then predict the risk values $R_{\text{first}}$ for the pair $\{V_{\text{ego}}, V_{\text{first}}\}$. If the predicted risk exceeds a predefined threshold $R_{\text{thr}}$, the situation is identified as high-risk and the interaction $\{V_{\text{ego}}, V_{\text{first}}\}$ is retrieved as a valuable driving situation. 

\begin{algorithm}[t]
\caption{Retrieving valuable first-order situations}\label{alg:algo_first_order}
\begin{algorithmic}[1]
\For{$V_{\text{ego}}$ and $V_{\text{first}}$ in vehicles}  
    \If{$V_{\text{first}} \neq V_{\text{ego}}$}    
        \State Predict risk $R_{\text{first}}$ for $V_{\text{ego}}$ and $V_{\text{first}}$
        \If{$R_{\text{first}} \geq R_{\text{thr}}$}
            \State Retrieve situation $\{V_{\text{ego}}, V_{\text{first}}\}$
        \EndIf
    \EndIf
\EndFor
\end{algorithmic}
\end{algorithm}

\begin{algorithm}[t]
\caption{Retrieving valuable second-order situations}\label{alg:algo_second_order}
\begin{algorithmic}[1]
\For{$V_{\text{ego}}$, $V_{\text{first}}$ and $V_{\text{second}}$ in vehicles}  
    \If{$V_{\text{first}} \neq V_{\text{ego}}$ \textbf{and} $V_{\text{second}} \notin \{V_{\text{ego}}, V_{\text{first}}\}$}    
        \State Predict risk $R_{\text{first}}$ for $V_{\text{ego}}$ and $V_{\text{first}}$
        \State Predict risk $R_{\text{second}}$ for $V_{\text{first}}$ and $V_{\text{second}}$
        \If{$R_{\text{first}} \geq R_{\text{thr}}$ \textbf{and} $R_{\text{second}} \geq R_{\text{thr}}$}
            \State Retrieve situation $\{V_{\text{ego}}, V_{\text{first}}, V_{\text{second}}\}$
        \EndIf
    \EndIf
\EndFor
\end{algorithmic}
\end{algorithm}

\subsection{Second-Order Driving Situations}
\label{subsec:second_order}
To retrieve second-order driving situations, we need to additionally predict risk values for a second vehicle pair. We can then trace back the risk flow in the interaction graph. The process of retrieving valuable second-order situations is given in Algorithm \ref{alg:algo_second_order}. 

For one data example, we iterate over all vehicles, selecting an ego vehicle $V_{\text{ego}}$, first other vehicle $V_{\text{first}}$ and a second other vehicle $V_{\text{second}}$. The first vehicle must be here different than the ego vehicle and the second vehicle must differ both from the ego and first vehicle. Then, we compute the risk $R_{\text{first}}$ for the first vehicle pair $\{V_{\text{ego}}, V_{\text{first}}\}$, and the risk $R_{\text{second}}$ for the second vehicle pair $\{V_{\text{ego}}, V_{\text{second}}\}$. Only if both risks $R_{\text{first}}$ and $R_{\text{second}}$ exceed the threshold $R_{\text{thr}}$, the situation is classified as high-risk, and the interaction involving $\{V_{\text{ego}}, V_{\text{first}}, V_{\text{second}}\}$ is retrieved as a valuable driving situation.

Using the proposed risk model and Algorithms \ref{alg:algo_first_order} and \ref{alg:algo_second_order}, we can identify valuable driving situations that enhance the quality of tests for software models in automated driving.

\section{Experimental Setup}
\label{sec:exp_setup}

To verify the risk-based filtering approach proposed in this paper, we apply our method to a large-scale driving dataset, the Waymo Open Motion Dataset \cite{waymo2021}. The dataset provides position and velocity information for each vehicle in recorded driving scenarios. We used the full pre-processed dataset and compute the risk values for all data and retrieved valuable driving situations. The dataset comprises over 500 hours of driving data collected across various U.S. cities, including San Francisco, Phoenix, Mountain View, Los Angeles, Detroit and Seattle. Although the evaluation in this work is based on the Waymo dataset, the risk model is general and can be applied to other datasets as well.

\subsection{Baselines}
In the experiments, we compare the proposed risk-based filtering approach with Kalman difficulty and Tracks-To-Predict (TTP). Kalman difficulty \cite{unitraj2024} determines the minimum final displacement error (minFDE) between a linear predicted trajectory and the ground truth. We compute the Kalman difficulty using a prediction horizon of $8\unit[]{s}$ and apply a threshold of $10 \unit[]{m}$ to identify valuable driving situations.

TTP is a heuristic used by Waymo in the Waymo dataset to identify interactive driving situations. It is based on a set of semantic rules, such as ``agent A changed lanes'', ``agent B crossed paths with agent C'', ``pedestrian D is close to vehicle E'' or ``agent F interacted with agent G with high acceleration''. However, the exact defintions of these rules and the associated thresholds are not publicly disclosed. In our evaluation, we use the provided TTP field in the dataset, which is $0$ for non-interesting and $1$ for interesting vehicles.

\subsection{Parameter Values}
The parameter values used in the risk model are summarized in Tab. \ref{tab:parameters}. The parameters were finetuned based on prior studies and datasets. The model assigns different maximum uncertainties to the longitudinal and lateral directions of motion, depending on the road user type. For example, cars can exhibit large longitudinal uncertainties up to $\sigma_{\text{car,max}} = 15\unit[]{m}$, while pedestrians have small lateral uncertainties, limited to $\sigma_{\text{ped,max}} = 1.5\unit[]{m}$. Bicycles have longitudinal uncertainties of up to $\sigma_{\text{cyc,max}} = 3.3\unit[]{m}$. Here, we define the starting uncertainties of the trajectories equal to the size of the road user.

In the risk integration, we set the avoidance rate parameter of the survival function to $\tau_0^{-1} = 0.56\hspace*{0.05cm}\unit[]{1/s}$ and the prediction timestep in the trajectory to $\Delta t=0.25\unit[]{s}$. Finally, the threshold of the risk value for detecting valuable driving situations with the risk-based filtering approach is set to $R_{\text{valuable}} = 10^{-9}$. 

\begin{table}[t!]
\fontsize{7.1}{8.1}\selectfont 

\centering
\resizebox{0.98\columnwidth}{!}{
\begin{tabular}{c c c} 
\toprule
Parameter & Description & Value \\
\midrule
$\sigma_{\text{car,max}}$ & longitudinal maximum uncertainty (cars)  $[\unit[]{m}]$ & $15$ \\ 
$\sigma_{\text{ped,max}}$ & lateral maximum uncertainty (pedestrians)  $[\unit[]{m}]$ & $ 1.5$ \\
$\sigma_{\text{cyc,max}}$ & longitudinal maximum uncertainty (bicycles) $[\unit[]{m}]$ & $3.3$ \\
$\tau_0^{-1}$ & avoidance rate $[\unit[]{1/s}]$ & $0.56$ \\
$s_{\text{max}}$ & maximum prediction time $[\unit[]{s}]$ & $8$ \\
$\Delta t$ & prediction timestep $[\unit[]{s}]$ & $0.25$ \\
$R_{\text{valuable}}$ & threshold for valuable driving situation & $10^{-9}$ \\
\bottomrule
\end{tabular}
}
\vspace{0.05cm}
\caption[]{Parameter values used in the risk model. The main parameters are the uncertainties for the vehicle types, including cars, pedestrians and bicycles.}
\vspace{-0.1cm}
\label{tab:parameters}
\end{table}

\subsection{Fine-Tuning Results}
To further improve the performance of the risk-based filtering, we made the following modifications aimed at fine-tuning the results and reducing noise in the driving data.

Trajectory prediction uncertainties can become large, particulary during turns, sometimes extending beyond the actual driving path. To address this, we employed a Gaussian mixture model to represent complex uncertainty shapes using multiple smaller components. This allows the overall uncertainty to follow the curvature of the road user's motion more accurately. For implementation details, we refer to \cite{puphal2019}.

Furthermore, we excluded situations in which both vehicles have speeds close to zero or follow only short driving paths. In such cases, the detected vehicles are often present in the data stream for a too short duration to yield reliable or meaningful trajectory information. With all of these parameter settings and refinements, our approach is able to effectively retrieve driving situations.

\vspace{-0.05cm}
\section{Results}
\label{sec:results}

This section presents the experimental results of the risk-based filtering approach applied to the Waymo dataset. The risk model is applied on each pair of vehicles in the full dataset. For each vehicle, we use here the position and velocity at timestep zero in the data stream and predict its future motion under constant velocity. The prediction is performed along the vehicle's driving path, which is extracted by analyzing the full recording of the scenario.

The section is divided into three parts. First, we compare the results of our method against the baselines of Kalman difficulty and TTP using a confusion matrix. Second, we provide statistics on the types of road users involved in valuable driving situations, including both first-order and second-order situations. Third, we present examples of driving situations that are seen valuable or non-valuable by the risk model. 

\subsection{Comparison with Baselines}

\begin{figure}[t!]
  \centering
  \vspace*{0.1cm}
  \resizebox{0.925\linewidth}{!}{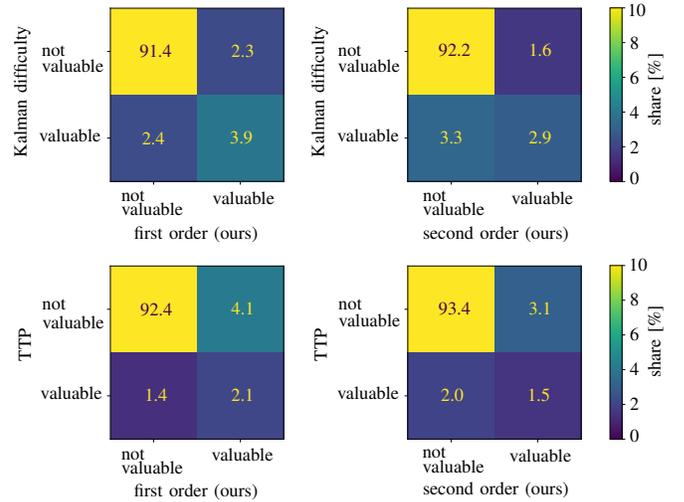}
  \caption[]{Confusion matrix comparing the valuable road users filtered from the Waymo Open Motion Dataset \cite{waymo2021}. We show the results of our risk model alongside those of Kalman difficulty and TTP.}
  \label{fig:confusion_matrix}
\end{figure}

Fig. \ref{fig:confusion_matrix} presents the confusion matrices comparing the sets of valuable road users identified by the different filtering approaches. The left side of the figure compares the first-order situations retrieved by our approach against those identified by Kalman difficulty and TTP, and the right side presents the comparison for second-order situations. Two main observations can be made. On the one hand, the majority of driving situations are considered as not valuable by all methods, as shown by the dominant yellow regions in the top-left corner of the matrices (exceeding $90 \unit[]{\%}$). On the other hand, many valuable first-order and second-order situations from our approach differ from those of the baselines. The off-diagonal entries of the matrix range between $1 \unit[]{\%}$ and $4 \unit[]{\%}$. 

As one example, out of a total of 67.8 million road users in the dataset, more than 1.5 million are detected as valuable by our approach but not by Kalman difficulty (see the $2.3 \unit[]{\%}$ entry in the matrix comparing first-order situations and Kalman difficulty in the top left). The differences are even larger when compared with TTP. However, this comparison is not exact, as we cannot recompute TTP. Driving data can repeat within the Waymo dataset, and the same vehicle may be flagged as valuable by TTP in one instance but not in another.

\subsection{Statistics of Valuable Driving Situation} 

\begin{figure}[t!]
  \centering
  \resizebox{0.975\linewidth}{!}{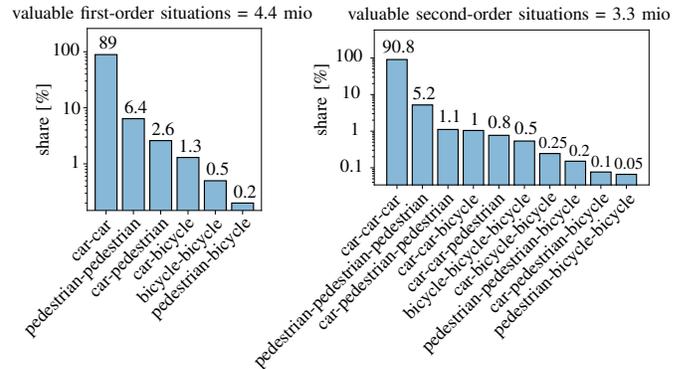}
  \vspace*{0.1cm}
  \caption[]{Histograms of road user types for filtered first-order and second-order valuable driving situations in the Waymo dataset.}
  \label{fig:second_order_risks}
\end{figure}

Fig. \ref{fig:second_order_risks} shows the histograms of the road user types involved in the valuable driving situations filtered by our approach. In the Waymo dataset, the detected road user types include cars, pedestrians and biycles. Most first-order situations (left image) involve car-car and pedestrian-pedestrian interactions. Car-car situations account for around $89 \unit[]{\%}$ of the cases and pedestrian-pedestrian situations for $6.4 \unit[]{\%}$. Bicycle-related interactions are less frequent, for example, car-bicycle situations represent only about $1.3 \unit[]{\%}$ of the total. 

For second-order situations (right image), the shares are similar. Car-car-car and pedestrian-pedestrian-pedestrian interactions are the most common, comprising $90.8 \unit[]{\%}$ and $5.2 \unit[]{\%}$ of cases, respectively. Bicycle-bicycle-bicycle situations are comparatively rare, representing only $0.5 \unit[]{\%}$ of the total.

\subsection{Example Driving Situations}
\begin{figure}[t!]
  \vspace*{0.17cm}
  \centering
  \resizebox{1.0\linewidth}{!}{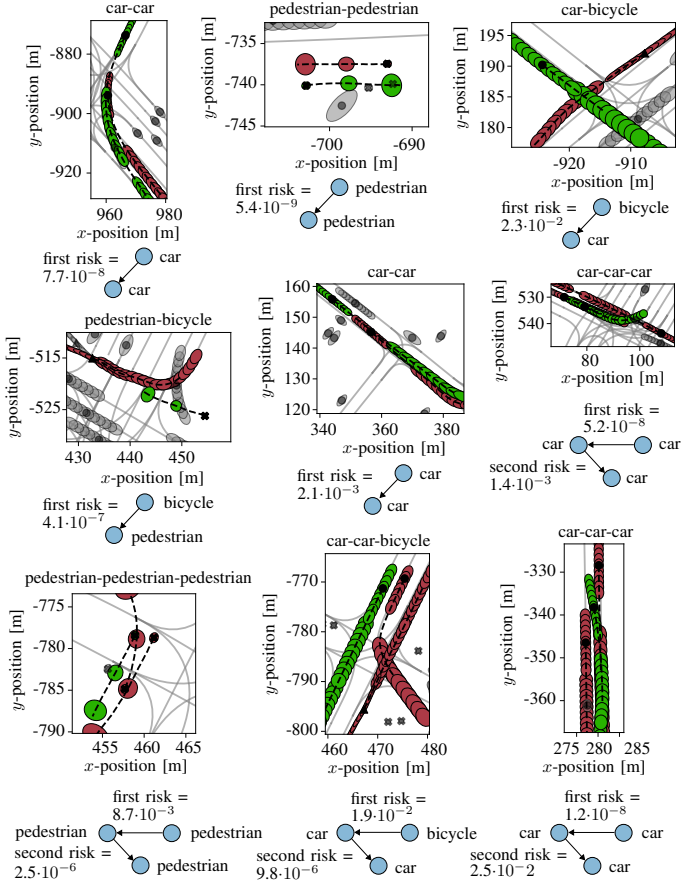}
  \vspace*{-0.27cm}
  \caption[]{Examples of valuable driving situations identified by the risk model in the Waymo dataset. The risk values are exceeding $R_{\text{valuable}} = 10^{-9}$ in the examples.} 
  \vspace*{-0.03cm}
  \label{fig:examples_interesting_situations}
\end{figure}

\begin{figure}[t!]
  \vspace*{0.05cm}
  \centering
  \resizebox{0.95\linewidth}{!}{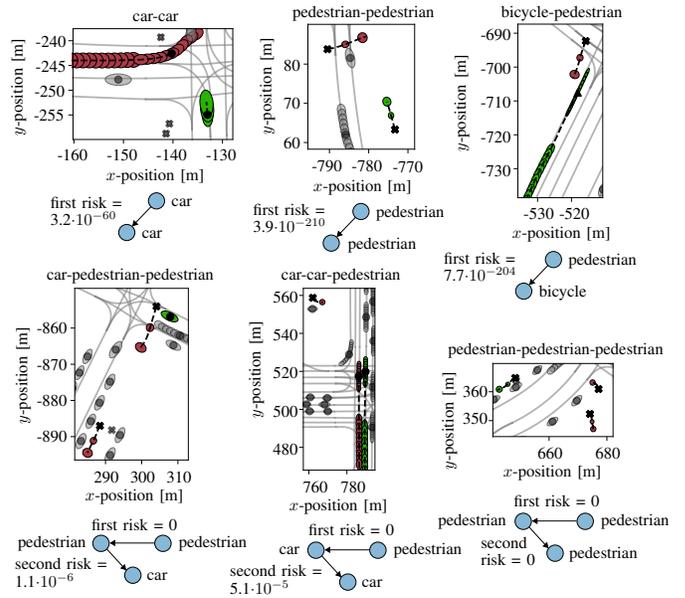}  
  \vspace*{-0.12cm}
  \caption[]{Examples of driving situations considered not valuable by the risk model in the Waymo dataset. In each example, at least one risk value is below $R_{\text{valuable}} = 10^{-9}$.} 
  \label{fig:examples_limiting_examples}
\end{figure}

Finally, example driving sitations identified as valuable by the risk model are given in Fig. \ref{fig:examples_interesting_situations}, and examples considered not valuable are shown in Fig. \ref{fig:examples_limiting_examples}. Each example depicts here the vehicle trajectories with their Gaussian uncertainties (top) as well as the corresponding interaction graph visualizing the risk values between the road users (bottom).

In valuable first-order driving situations, the risk between pairs of road users exceeds $R_{\text{valuable}} = 10^{-9}$ as the threshold. Examples of such situations, as illustrated in Fig. \ref{fig:examples_interesting_situations}, include close car-car following situations, close pedestrian-pedestrian passing situations, and crossing situations involving cars and bicycles or pedestrians and bicycles. Second-order situations can include, e.g., lane-cutting maneuvers involving two other cars, three pedestrians moving in a crowd, bicycles turning in multi-lane traffic or complex car merging scenarios.   

Example driving situations which are not valuable based on the risk model have one risk value in the driving situation that is below $R_{\text{valuable}} = 10^{-9}$. Fig. \ref{fig:examples_limiting_examples} shows such examples of non-valuable first-order situations, e.g., car-car situations with one vehicle waiting and not interacting, pedestrian-pedestrian situations with large distances and bicycle-pedestrian situations in which the interaction already happened. For second-order situations, the figure shows group interactions in which one road user is far away, or vehicles that are not interacting in the near future. 

\section{Conclusion and Outlook}
\label{sec:conclusion}

In summary, in this paper, we presented the application of the proposed risk-based filtering approach on the Waymo Open Motion Dataset \cite{waymo2021}. By applying the risk model to each road user in the dataset and creating interaction graphs with risk labels, we were able to classify and retrieve two types of valuable driving situations. The driving situation types were: a) first-order situations (where one vehicle directly influences another and induces risk) and b) second-order situations (where influence propagates through an intermediary vehicle). 

Finally, we compared our approach against baseline methods. The experiments show that our risk model is capable of retrieving complementary situations not captured by the existing methods of Kalman difficulty and Tracks-To-Predict (TTP). This capability can contribute to the development of more robust, machine learning-based trajectory prediction models and improve testing of automated vehicles in complex conditions. The risk values for the Waymo dataset, along with the IDs of the valuable driving situations, are open-source and available upon request. The Github repository can be found at \url{https://github.com/HRI-EU/RiskBasedFiltering}. 

As future work, we plan to extend the application of our risk model to other datasets and release additional valuable driving situations to the research community. In particular, datasets such as Nuscenes dataset \cite{nuscenes2020} and Argoverse dataset \cite{argoverse2021}, which currently also rely on handcrafted filtering, could benefit from a unified and risk-driven filtering approach. Furthermore, we aim to use the retrieved valuable driving situations to train a machine learning-based trajectory predictor that performs well under risky driving conditions. 

\addtolength{\textheight}{-12cm}   

\vspace{0.13cm}
\bibliographystyle{IEEEtran}
\nopagebreak
\bibliography{bib}

\begin{thebibliography}{10}
\providecommand{\url}[1]{#1}
\csname url@samestyle\endcsname
\providecommand{\newblock}{\relax}
\providecommand{\bibinfo}[2]{#2}
\providecommand{\BIBentrySTDinterwordspacing}{\spaceskip=0pt\relax}
\providecommand{\BIBentryALTinterwordstretchfactor}{4}
\providecommand{\BIBentryALTinterwordspacing}{\spaceskip=\fontdimen2\font plus
\BIBentryALTinterwordstretchfactor\fontdimen3\font minus
  \fontdimen4\font\relax}
\providecommand{\BIBforeignlanguage}[2]{{%
\expandafter\ifx\csname l@#1\endcsname\relax
\typeout{** WARNING: IEEEtran.bst: No hyphenation pattern has been}%
\typeout{** loaded for the language `#1'. Using the pattern for}%
\typeout{** the default language instead.}%
\else
\language=\csname l@#1\endcsname
\fi
#2}}
\providecommand{\BIBdecl}{\relax}
\BIBdecl

\bibitem{waymo2021}
S.~Ettinger, S.~Cheng, and et~al., ``{Large Scale Interactive Motion
  Forecasting for Autonomous Driving: The Waymo Open Motion Dataset},'' in
  \emph{IEEE/CVF International Conference on Computer Vision (CVPR)}, 2021.

\bibitem{nuscenes2020}
H.~Caesar, V.~Bankiti, A.~H. Lang, and et~al., ``{nuScenes: A Multimodal
  Dataset for Autonomous Driving},'' in \emph{IEEE/CVF Conference on Computer
  Vision and Pattern Recognition (CVPR)}, 2020.

\bibitem{argoverse2021}
B.~Wilson, W.~Qi, T.~Agarwal, and et~al., ``{Argoverse 2: Next Generation
  Datasets for Self-Driving Perception and Forecasting},'' in \emph{Conference
  on Neural Information Processing Systems (NeurIPS)}, 2021.

\bibitem{puphal2019}
T.~Puphal, M.~Probst, and J.~Eggert, ``{Probabilistic Uncertainty-Aware Risk
  Spot Detector for Naturalistic Driving},'' \emph{IEEE Transactions on
  Intelligent Vehicles (T-IV)}, 2019.

\bibitem{unitraj2024}
L.~Feng, M.~Bahari, K.~M.~B. Amor, and et~al., ``{UniTraj: A Unified Framework
  for Scalable Vehicle Trajectory Prediction},'' in \emph{European Conference
  on Computer Vision (ECCV)}, 2024.

\bibitem{tolstaya2021}
E.~Tolstaya, R.~Mahjourian, C.~Downey, and et~al., ``{Identifying Driver
  Interactions via Conditional Behavior Prediction},'' in \emph{IEEE
  International Conference on Robotics and Automation (ICRA)}, 2021.

\bibitem{mavrogiannis2022}
C.~Mavrogiannis, J.~DeCastro, and S.~Srinivasa, ``{Analyzing Multiagent
  Interactions in Traffic Scenes via Topological Braids},'' in \emph{IEEE
  International Conference on Robotics and Automation (ICRA)}, 2022.

\bibitem{klingelschmitt2016}
S.~Klingelschmitt and et~al., ``{Probabilistic Situation Assessment Framework
  for Multiple, Interacting Traffic Participants in Generic Traffic Scenes},''
  in \emph{IEEE Intelligent Vehicles Symposium (IV)}, 2016.

\newpage

\bibitem{RSS2017}
S.~Shalev{-}Shwartz, S.~Shammah, and A.~Shashua, ``{On a Formal Model of Safe
  and Scalable Self-driving Cars},'' \emph{arXiv:1708.06374}, 2017.

\bibitem{MIT2022}
W.~Han, A.~Jasour, and B.~Williams, ``{Non-Gaussian Risk Bounded Trajectory
  Optimization for Stochastic Nonlinear Systems in Uncertain Environments},''
  in \emph{IEEE International Conference on Robotics and Automation (ICRA)},
  2022.

\bibitem{eggert2017}
J.~Eggert and T.~Puphal, ``{Continuous Risk Measures for ADAS and AD},''
  \emph{Future Active Safety Technology towards Zero-Traffic-Accidents
  (FAST-zero)}, 2017.

\bibitem{interaction2019}
W.~Zhan, L.~Sun, and et~al., ``{INTERACTION Dataset: An INTERnational,
  Adversarial and Cooperative moTION Dataset in Interactive Driving Scenarios
  with Semantic Maps},'' \emph{arXiv:1910.03088}, 2019.

\end{thebibliography}
\end{document}